\documentclass[journal]{IEEEtran}

\usepackage{my}

\newcommand{\imdbwiki}{IMDB-WIKI}
\newcommand{\imdbc}{IMDB-Clean}
\newcommand{\morph}{Morph}
\newcommand{\kanface}{KANFace}
\newcommand{\fgnet}{FG-Net}
\newcommand{\cacd}{CACD}
\newcommand{\fpage}{FP-Age}

\begin{document}
\title{{\fpage}: Leveraging Face Parsing Attention \\ for Facial Age Estimation in the Wild}
\author{Yiming~Lin,~\IEEEmembership{Member,~IEEE,}
        Jie~Shen,~\IEEEmembership{Member,~IEEE,}
        Yujiang~Wang,
        and~Maja~Pantic,~\IEEEmembership{Fellow,~IEEE}%
\thanks{Yiming Lin, Jie Shen, Yujiang Wang and Maja Pantic are with the Department of Computing, Imperial College London, UK (e-mail:  {yiming.lin15@imperial.ac.uk}; {jie.shen07@imperial.ac.uk}; {yujiang.wang14@imperial.ac.uk}; {maja.pantic@gmail.com}).
}%
\thanks{Jie Shen is the corresponding author (e-mail: jie.shen07@imperial.ac.uk).}%
\thanks{Code is available at \url{https://github.com/ibug-group/fpage}.}%
\\
\url{https://github.com/ibug-group/fpage}
}

\markboth{IEEE TRANSACTIONS ON IMAGE PROCESSING,~Vol.~xx, No.~x, xx~xxxx}%
{Lin \MakeLowercase{\textit{et al.}}: FP-Age: Leveraging Face Parsing Attention for Facial Age Estimation in the Wild}

\maketitle

\begin{abstract}
Image-based age estimation aims to predict a person's age from facial images. It is used in a variety of real-world applications. Although end-to-end deep models have achieved impressive results for age estimation on benchmark datasets, their performance in-the-wild still leaves much room for improvement due to the  challenges caused by large variations in head pose, facial expressions, and occlusions. To address this issue, we propose a simple yet effective method to explicitly incorporate facial semantics into age estimation, so that the model would learn to correctly focus on the most informative facial components from unaligned facial images regardless of head pose and non-rigid deformation. To this end, we design a face parsing-based network to learn semantic information at different scales and a novel face parsing attention module to leverage these semantic features for age estimation. To evaluate our method on in-the-wild data, we also introduce a new challenging large-scale benchmark called IMDB-Clean. This dataset is created by semi-automatically cleaning the noisy IMDB-WIKI dataset using a constrained clustering method. Through comprehensive experiment on IMDB-Clean and other benchmark datasets, under both intra-dataset and cross-dataset evaluation protocols, we show that our method consistently outperforms all existing age estimation methods and achieves a new state-of-the-art performance. To the best of our knowledge, our work presents the first attempt of leveraging face parsing attention to achieve semantic-aware age estimation, which may be inspiring to other high level facial analysis tasks.

\end{abstract}

\begin{IEEEkeywords}
Age estimation, face parsing, in-the-wild dataset, attention, cross-dataset evaluation.
\end{IEEEkeywords}

\IEEEpeerreviewmaketitle

\section{Introduction}
\IEEEPARstart{A}{ge} estimation from facial images has been an active research topic in computer vision and it can be utilised in a variety of real-world applications, such as forensics, security, health and well-being, and social media. There are several branches in this topic. In this work, we focus on the estimation of real/biological age, which is arguably the most difficult task among others such as apparent age estimation~\cite{Escalera_2016_CVPR_Workshops} or age group classification~\cite{Levi_2015_CVPR_Workshops}. Predicting a person's age from facial images in the wild can be very challenging as it involves a variety of intrinsic and subtle factors such as pose, expression, gender, illuminations, occlusions, \etc

Recently, deep learning approaches have been widely employed to construct end-to-end age estimations models. Deep embedding learnt from large-scale datasets is a very effective facial representation that has greatly improved the state-of-the-art in automatic estimation of facial age. However, most deep models are not explicitly trained to learn facial semantic information like eyes and noses, and therefore the extracted embedding may not appropriately attend to those more informative facial regions.

It has been shown that the most informative features for age estimation are located in the local regions such as eyes and mouth corners~\cite{hanDemographicEstimationFace2015}. On the other hand, face parsing is designed to classify each pixel into different facial regions and to give the regional boundaries. Therefore, a Convolutional Neural Network~(CNN) trained for face parsing could also pick up the features around the facial regions that are also useful for determining the age. Moreover, due to the hierarchical structure of CNNs, the intermediate features can encode both local and global information that can be fused for age estimation.

To this end, we propose FP-Age for leveraging features in a face parsing network for facial age estimation. 
In particular, we adopt both coarse and fine-grained features from a pre-trained face parsing network \cite{linRoITanhpolarTransformer2021} to represent facial semantic information at different levels and built a small network on top of it to predict the age.
To avoid the loss of details in the high level features, we design a Face Parsing Attention (FPA) module to explicitly drive the network's attention to those more informative facial parts. 
The attended high-level features are then concatenated to the low-level features and fed into a small add-on network for age prediction.
Since the semantic features are extracted using a pre-trained face parsing model, no additional face parsing annotations are required and thus our \fpage~network can be trained in an end-to-end fashion, similar to other age estimation networks.

We have also developed a semi-automatic approach to clean the noisy data in \imdbwiki, leading to a new large-scale age estimation benchmark titled \imdbc. 
Our FP-Age network achieves state-of-the-art results on this \imdbc~, as well as on several other age estimation datasets, under both intra-dataset and cross-dataset evaluation protocols. 
To the best of our knowledge, this is the first reported effort to adopt semantic facial information for age estimation based on an attention mechanism on different facial regions.
The idea of Face Parsing Attention can be inspiring to other facial analysis tasks too, and the proposed FP-Age network can be easily adapted to perform on those tasks as well, \eg facial gesture recognition and emotion recognition.

Our main contributions are as follows:
\begin{itemize}
    \item The \imdbc~dataset: a large-scale, clean image dataset for age estimation in the wild;
    \item \fpage: a simple yet effective 
    framework that leverages facial semantic features for semantic-aware age estimation;
    \item We also demonstrate that for age estimation, different facial parts have variable importance with ``nose'' being the least important region;
    \item Our \fpage~achieves new state-of-the-art results on \imdbc, \morph~\cite{ricanekMORPHLongitudinalImage2006} and \cacd~\cite{chenFaceRecognitionRetrieval2015};
    \item When trained on \imdbc, our \fpage~also achieves state-of-the-art results on \kanface~\cite{georgopoulosInvestigatingBiasDeep2020}, \fgnet~\cite{fgnet2002}, \morph~\cite{ricanekMORPHLongitudinalImage2006} and \cacd~\cite{chenFaceRecognitionRetrieval2015} under cross-dataset evaluation.
\end{itemize}
\begin{figure*}[!t]
    \centering
    \includegraphics[width=\textwidth]{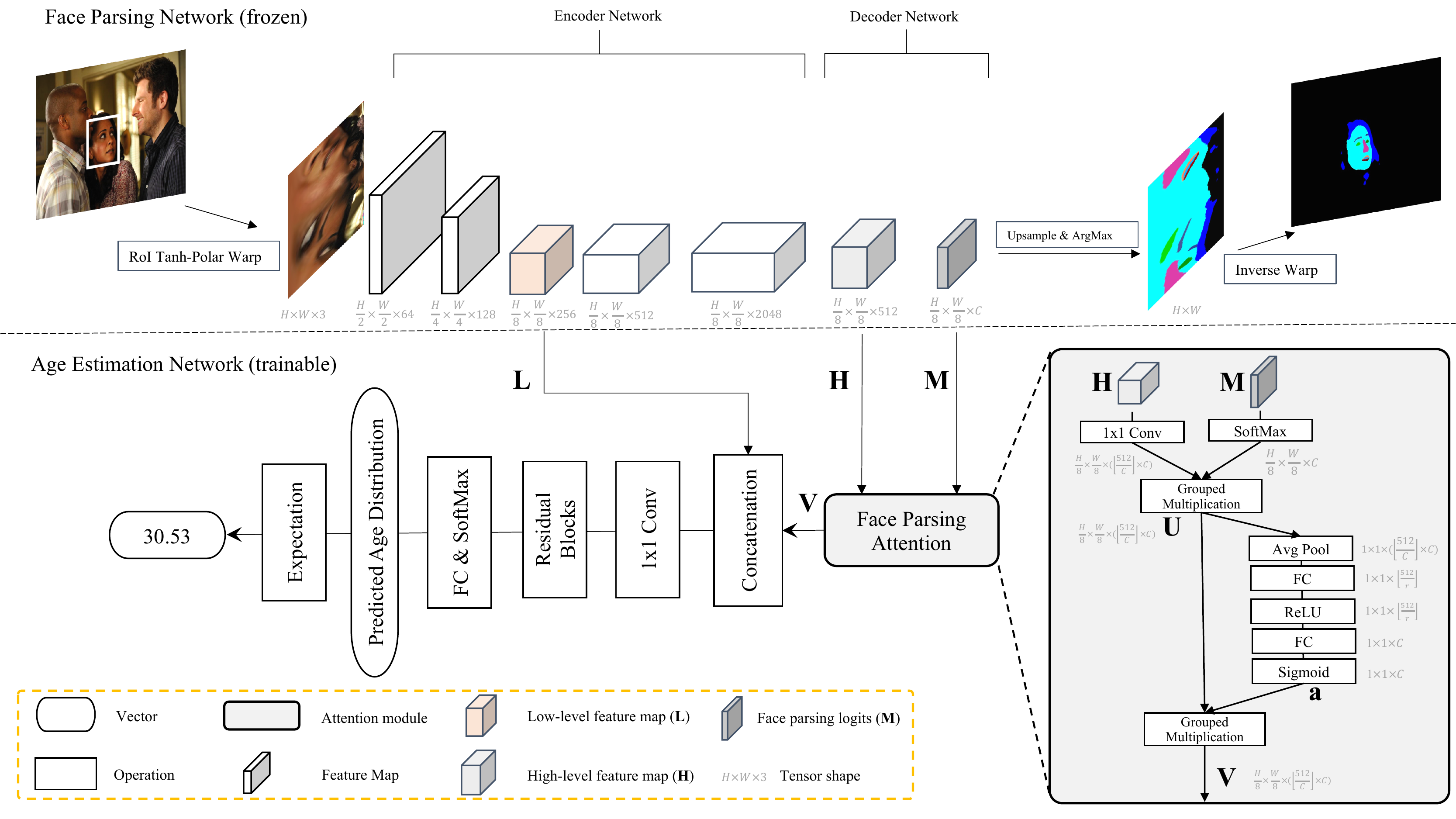}
    \caption{FP-Age. A pre-trained face parsing framework~\cite{linRoITanhpolarTransformer2021} (top) is used to extract features of the target face in the input image. A lightweight network (bottom) aggregates low-level features, high-level features and face masks to predict the age. The shapes of tensors are labelled by the blocks and $\lfloor \cdot \rfloor$ means floor division. Face Parsing Attention is proposed to aggregate the semantic information into the features and improve age estimation.}
    \label{fig:overall}
\end{figure*}
\section{Related Work}\label{sec:related-work}

\subsection{Image-based Biological Age Estimation}
Early works on age estimations are mainly based on handcrafted features, and we refer interested readers to \cite{panisOverviewResearchFacial2016} for a detailed survey. Recently, deep learning techniques have achieved significantly improved performance in this field.  
In this section, we briefly explain several deep learning approaches on age estimation. They are roughly organised into four categories depending on how they model the problem: regression based, classification based, ranking based and label distribution based. 

Regression approaches treat facial ageing as a regression problem and directly predict true age values from facial images. Euclidean loss is therefore a popular choice among those methods.   
Yi~\etal \cite{yiAgeEstimationMultiscale2014} adopted mean squared loss to train a multi-scale CNN for age regression.  Similarly, Wang~\etal~\cite{wangDeeplyLearnedFeatureAge2015} apply the same loss on the representation obtained by fusing feature maps from different layers of a CNN. 

In contrast to regression methods, classification based works~\cite{Levi_2015_CVPR_Workshops, EidingerAge2014} formulate the age estimation as a multi-class classification problem and treat different ages as independent classes. 
Although such formulations make it easier to train CNNs, this ignores the correlations between different classes.

Ranking approaches inspect the ordinal property embedded in the ageing process. OR-CNN \cite{niuOrdinalRegressionMultiple2016} proposed to formulate age estimation as an ordinal regression problem and built multiple binary classification neurons on top of a CNN. 
Ranking-CNN~\cite{chenUsingRankingCNNAge2017} ensembled a series of CNN-based binary classifiers and aggregated their predictions to obtain the estimated age. In SVRT~\cite{imScaleVaryingTripletRanking2019}, a strategy of  triplet learning were introduced into the ranking loss.
CORAL~\cite{Cao2020coral} improved OR-CNN \cite{niuOrdinalRegressionMultiple2016} by proposing the Consistent Rank Logits framework to address the problem of classifier inconsistency.

Label Distribution Learning (LDL) \cite{gengLabelDistributionLearning2016}, however, models the age prediction as a probability distribution over all potential age values. LDL-based methods have achieved the current state-of-the-art performance on various age estimation benchmarks. Dex~\cite{rotheDEXDeepEXpectation2015, rotheDeepExpectationReal2018} proposed to take the expectation value of output distribution as the predicted age. MV-Loss~\cite{panMeanVarianceLossDeep2018} introduced the mean–variance loss to regularise the shape of the output distribution complementing the cross-entropy loss. DLDL~\cite{gaoDeepLabelDistribution2017} and DLDL-v2~\cite{gaoAgeEstimationUsing2018} represented the age label as a Gaussian distribution and applied Kullback-Leibler divergence to measure the discrepancy between the output age distribution and the target label distribution. Shen~\etal\cite{shenDeepRegressionForests2018, shenDeepDifferentiableRandom2021} used an ensemble of decision trees in the LDL formulation. Akbari~\etal\cite{akbariDistributionCognisantLoss2020} proposed the distribution cognisant loss to regularise the predicted age distribution, improving the robustness against outliers. In this work, we follow the problem formulation of LDL-based methods, considering that they have consistently achieved most state-of-the-art results.

Noticeably, several approaches~\cite{ranjanAllInOneConvolutionalNeural2017, imScaleVaryingTripletRanking2019, gaoAgeEstimationUsing2018} involve applying pre-trained face recognition models as the initialisation of ages estimation models, while in contrast, we freeze the weights of the face parsing network to avoid unnecessary computational cost. 
Additionally, some works~\cite{Levi_2015_CVPR_Workshops, hanHeterogeneousFaceAttribute2018, wangDeepMultiTaskLearning2017, ranjanAllInOneConvolutionalNeural2017} tackled age estimation simultaneously with other tasks like gender classification through a multi-task framework, sharing representations across different tasks. 
Although our network also share features, it differs from multi-task framework as it requires no semantic labels and also Face Parsing Attention is leveraged to transit semantic-level knowledge.
\subsection{Face Parsing}
Face parsing aims to classify each pixel in a facial image into different categories like background, hair, eyes, nose, \etc. Earlier works \cite{Warrell2009,Smith_2013_CVPR} used holistic priors and hand-crafted features. 
Deep learning has largely improved the performance of face parsing models
Liu~\etal\cite{liu2015multi} combined CNNs with conditional random fields and proposed a multi-objective learning method to model pixel-wise likelihoods and label dependencies. 
Luo~\etal~\cite{luo2012hier} applied multiple Deep Belief Networks to detect facial parts and built a hierarchical face parsing framework. Jackson~\etal~\cite{jackson2016cnn} employed facial landmarks as a shape constraint to guide Fully Convolution Networks~(FCNs) for face parsing. 
Multiple deep methods including CRFs, Recurrent Neural Networks (RNNs) and Generative Adversarial Networks (GAN) were integrated by authors of \cite{gucclu2017end} to formulate an end-to-end trainable face parsing model, while the facial landmarks also served as the shape constraints for segmentation predictions. The idea of leveraging shape priors to regularise segmentation masks can also be found in the Shape Constrained Network (SCN) \cite{luo2020shape} for eye segmentation.
In~\cite{Liu2017Face}, a spatial Recurrent Neural Networks was used to model spatial relations within face segmentation masks. 
A spatial consensus learning technique was explored in \cite{masi2020structure} to model the relations between output pixels, while graph models were adopted in \cite{te2020edge} to learn implicit relationships between facial components. 
To better utilise the temporal information of sequential data, authors of \cite{wang2019face} integrated ConvLSTM \cite{xingjian2015convolutional} with the FCN model \cite{FCNs} to simultaneously learn the spatial-temporal information in face videos and to obtain temporally-smoothed face masks. 
In \cite{Wang_2020_CVPR}, a Reinforcement-Learning-based key scheduler was introduced to select online key frames for video face segmentation such that the overall efficiency can be globally optimised.

Most of those methods assume the target face has already been cropped out and is well aligned. Moreover, they often ignore the hair class due to the unpredictable margins for cropping the hair region. To solve this, Lin~\etal\cite{Lin_2019_CVPR} proposed to warp the entire image using the Tanh function. However, the warping still requires not only the facial bounding boxes but also the facial landmarks. Recently, RoI Tanh-polar transform~\cite{linRoITanhpolarTransformer2021} has been proposed to solve face parsing in the wild. The RoI Tanh-polar transform warps the entire image to the Tanh-polar space and the only requirement is to have the target bounding box. With the Tanh-polar representation, a simple FCN architecture has already achieved state-of-the-art results~\cite{linRoITanhpolarTransformer2021}. The proposed \fpage~builds atop of this method.

\section{Methodology}\label{sec:method}
The overall architecture of \fpage~is shown in \figurename~\ref{fig:overall}. The network at the top is an off-the-shelf, pre-trained face parsing model~\cite{linRoITanhpolarTransformer2021} whose parameters are not updated for the training. At the bottom is the proposed age estimation network that contains the proposed face parsing attention module and some standard operational layers to predict the age. 
In this section, we formulate age estimation as the distribution learning problem and explain further in detail the components in the proposed \fpage.

\subsection{Problem Formulation}
Let ${X} = \{(\mathbf{x}^{(i)},\mathbf b^{(i)},y^{(i)})\}_{i=1}^N$ denote a set of $N$ training example triplets where $\mathbf{x}^{(i)},  \mathbf{b}^{(i)}\text{ and } y^{(i)}$ are $i$-th input image, its target face bounding box, and its corresponding age label, respectively. The bounding box $\mathbf{b}^{(i)}$ is a four-dimensional tuple $(x_{min}, y_{min}, x_{max}, y_{max})$ defined by the top-left and the bottom-right corners of the target face location. The age label ${y^{(i)}}$ is an integer from a set of age labels $Y=\{0,	\dots, K-1\}$. We denote the total number of age classes Y as $K$. 

Our goal is to learn a mapping function $f$ from the target face in $\mathbf x^{(i)}$, specified by $\mathbf b^{(i)}$, to the label $y^{(i)}$. When learning such function using DNNs, one way is to set the last layer as one output neuron and employ an Euclidean loss function. However, it has been shown~\cite{rotheDEXDeepEXpectation2015, gaoDeepLabelDistribution2017} that training such DNNs are relatively unstable; outliers can cause large errors.
Another way is to formulate age estimation as a $K$-class classification  problem and use the one-hot encoding to represent age labels. But this formulation ignores the fact that the faces with close ages share similar features, causing visual label ambiguity~\cite{gengLabelDistributionLearning2016}.

Considering the above, we formulate the age estimation as a label distribution learning problem~\cite{gengLabelDistributionLearning2016}. Specifically, we encode each scalar age label $y^{(i)}$ as a probability distribution $\mathbf{q}^{(i)}=[q^{(i)}_0, q^{(i)}_1, ..., q^{(i)}_{K-1}]^T\in \mathcal{R}^K$ on the interval $[0, K-1]$. Each element in $\mathbf q^{(i)}$ represents the probability of the target face in $\mathbf x^{(i)}$ having the $k$-th label. A Gaussian distribution centred at $y^{(i)}$ with a standard deviation $\sigma$ is used to map $y^{(i)}$ to $\mathbf{q}^{(i)}$. We follow Gao \etal \cite{gaoAgeEstimationUsing2018} and set $\sigma=2$ in all experiments.

Using this formulation,  we use a Fully-Connected~(FC) layer followed by a Softmax layer to map the DNN's output logits to the predicted distribution $\mathbf p^{(i)}$. The learning problem becomes
\begin{equation}
    \theta^*=\argmin_{\theta}\sum_{i=1}^N L^{(i)} [\mathbf p^{(i)}=f(\mathbf x^{(i)}, \mathbf b^{(i)}), \mathbf{q}^{(i)}]
\end{equation}
where $f$ is the DNN and $\theta$ is its corresponding parameters. $L$ denotes a loss function. The predicted age $\hat{y}^{(i)}$ is obtained by taking the expectation over the $\mathbf p^{(i)}$ as $\hat{y}^{(i)}=\sum_{k=0}^{K-1}kp^{(i)}_k$.

\subsection{Face Parsing Network}
We use RTNet~\cite{linRoITanhpolarTransformer2021} for extracting face parsing features. RTNet has a simple FCN-like encoder-decoder architecture and achieves state-of-the-art results for in-the-wild face parsing tasks. The encoder contains $5$ residual convolutional layers for feature extraction, similar to the original ResNet-50~\cite{resnet}. Two convolutional layers are used in the decoder to perform per-pixel classification to obtain the face masks. In the encoder, the first three convolutional layers gradually reduce the spatial resolution to $({\frac{H}{8}, \frac{W}{8}})$, and the last two layers use dilated convolutions~\cite{YuKoltun2016} to aggregate multi-scale contextual information without reducing the resolution. 

\begin{figure}
    \centering
    \includegraphics[width=\columnwidth]{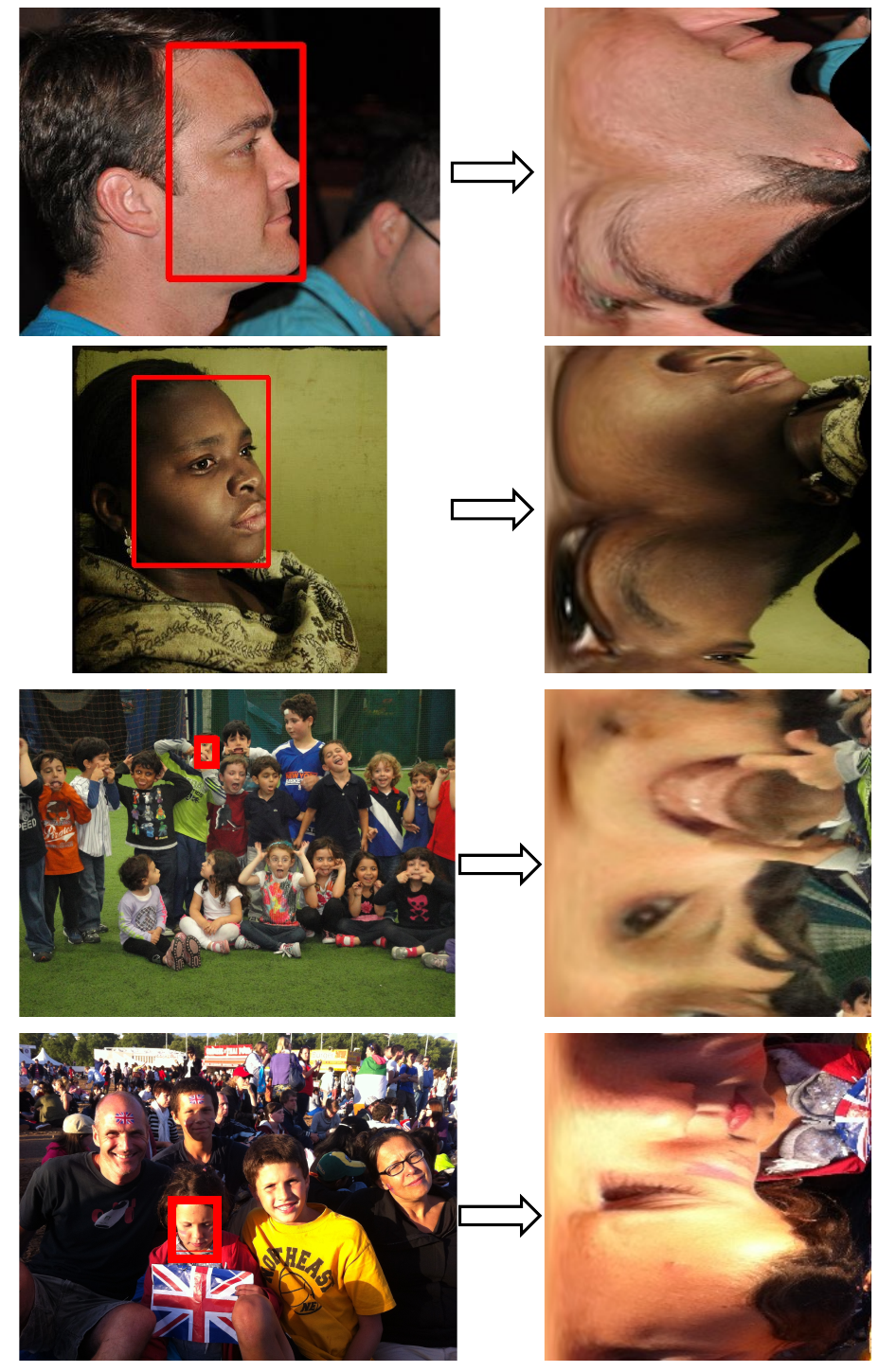}
    \caption{RoI Tanh-polar Transform~\cite{linRoITanhpolarTransformer2021} warps the whole image to a fix-sized representation in the Tanh-polar space with the given bounding box.}
    \label{fig:warping-results}
\end{figure}

In contrary to traditional methods that require facial landmarks to align the faces, RTNet uses RoI Tanh-polar transform to warp the entire image given the target bounding box. Some examples of the warping effect can be seen in \figurename~\ref{fig:warping-results}. The warped representation not only retains all the information in the original image, but also amplifies the target face.  

\subsection{Face Parsing Attention}
As shown in \figurename~\ref{fig:overall}, there are five feature maps produced by the encoder and one feature map given by the decoder. We take the third feature map in the encoder and denote it as the low-level feature $\mathbf{L}\in \mathbb{R}^{\frac{H}{8}\times \frac{W}{8}\times 256}$. We consider the only feature map in the decoder as the high-level feature and denote it as $\mathbf{H}\in \mathbb{R}^{\frac{H}{8}\times \frac{W}{8}\times 512}$. Lastly, we denote the output $C$-channel face masks as $\mathbf{M}\in \mathbb{R}^{\frac{H}{8}\times \frac{W}{8}\times C}$. 

We first divide $\mathbf{H}$ into $C$ groups along the channel dimension. The $k$-th is group representation, after $1\times 1$ convolution, is denoted as $\mathbf U_k\in \mathbb{R}^{\frac{H}{8}\times \frac{W}{8}\times \lfloor \frac{512}{C}	\rfloor	}$ for $k=1, \dots, C$. Next, we multiply each group with the corresponding mask group:

\begin{equation}
    \hat{\mathbf U}_k = \mathbf{M}{_k} * \mathbf{U}{_k} .
\end{equation}

The representations ${\hat{\mathbf U}_k} \text{ for } k = 1 \dots C$ are then concatenated along the channel dimension to get $U$. After that, we apply a channel attention block~\cite{Hu_2018_CVPR} to capture the dependencies between face regions. This block is formed by a sequence of layers: AvgPool, FC, ReLU, FC and Sigmoid. And the output attention weights are $\mathbf a\in \mathbb{R}^{C}$.
The final output of this module is $\mathbf V=[\mathbf V_1, \mathbf V_2, \dots, \mathbf V_{C}]$ and each feature group is obtained by
\begin{equation}
    \mathbf V_{k} = a_k \hat{\mathbf U}_{k}.
\end{equation}

\subsection{Age Estimation Network}
After the face parsing attention module is applied, we concatenate $\mathbf{L}$ and $\mathbf{V}$ along the channel dimension, and apply a $1\times 1$ convolutional layer to reduce the channel number to $256$. Next, $4$ residual blocks~\cite{resnet} are employed. Finally, we use a FC layer followed by a SoftMax layer to map the output logits to the predicted distribution $\mathbf p$. The predicted age is obtained by taking the expectation over $\mathbf p$ as $\hat{y}=\sum_{k=0}^{K-1}kp_k$.

.

\subsection{Loss Function}
We use the weighted sum of Kullback–Leibler divergence and L1 loss as our loss function for the $i$-th example: 
\begin{equation}
    L^{(i)} = \sum_{k=0}^{K-1}q^{(i)}_{k} log (\frac{q_k^{(i)}}{p_k^{(i)}}) + \lambda |\hat y^{(i)} - y^{(i)}|
\end{equation}
where $|\cdot|$ denotes taking the absolute value and $\lambda$ is a weight balancing two terms. We empirically set $\lambda=1$ for all examples~\cite{gaoDeepLabelDistribution2017}.

\begin{figure*}[t]
    \includegraphics[width=\textwidth]{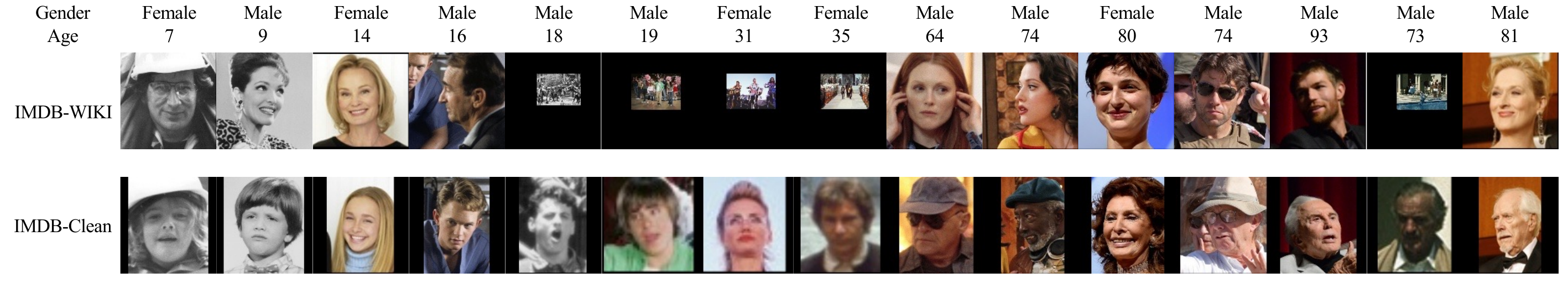}
    \caption{Some examples from IMDB-WIKI~\cite{rotheDEXDeepEXpectation2015} and our proposed IMDB-Clean. Each column shows the faces cropped from the same image using the groundtruth bounding boxes. The face detector used by IMDB-WIKI is biased towards middle-aged faces when encountering multiple faces, and fails for low-quality images. Our proposed semi-automatic cleaning method has corrected these errors (see Section~\ref{sec:creating-imdb-clean} for details).}
    \label{fig:compare-imdb-clean}
\end{figure*}

\section{Experimental Setup}\label{sec:exp}
\subsection{Exisiting Datasets}
\subsubsection{IMDB-WIKI}
The IMDB-WIKI~\cite{rotheDEXDeepEXpectation2015} is a large-scale dataset containing 523,051 images with age labels ranging from 0 to 100 years old. The images were crawled from IMDB and Wikipedia, where the IMDB subset contains $460,723$ images and the Wikipedia subset contains $62,328$ images. These images, especially the IMDB subset, were mostly captured in-the-wild and thus are potentially useful for evaluating age estimation in real-world environment.
However, the annotations of IMDB-Wiki are very noisy, such that the provided face box is often centred around the wrong person when multiple people are presented in the same image. Because of this, IMDB-Wiki has only been used for pre-training by existing age estimation methods~\cite{rotheDeepExpectationReal2018, zhangC3AEExploringLimits2019, panMeanVarianceLossDeep2018}.

\subsubsection{CACD}
Cross-Age Celebrity Dataset (CACD)~\cite{chenFaceRecognitionRetrieval2015} is an in-the-wild dataset that has about $160,000$ facial images of $2,000$ people. These images are divided into the training set, the validation set and the test set which
contain $1,800$ people, $120$ people and $80$ people, respectively. We adopt the common practice originally used in ~\cite{shenDeepDifferentiableRandom2021} and report results on the testing set obtained by using the models trained on the training set and the validation set.

\subsubsection{KANFace} KANFace~\cite{georgopoulosInvestigatingBiasDeep2020} is an in-the-wild dataset consisting of $41,036$ images from $1,045$ subjects. The age range of this dataset is from 0 to 100 years. The images are extremely challenging due to large variations in pose, expression and lightning conditions. Since the authors do not provide splits, we use this dataset only as a test set and the evaluation results obtained by models trained on other datasets.

\subsubsection{Morph} Morph~\cite{ricanekMORPHLongitudinalImage2006} consists of $55,134$ mugshot images from $13,617$ subjects with the age ranging from $16$ to $77$ years old. Even though it is not an in-the-wild dataset, we report our results on it given its popularity. For intra-dataset evaluations, we follow the setting used in~\cite{liuOrdinalDeepFeature2017, gaoAgeEstimationUsing2018, dengPMLProgressiveMargin2021}: we randomly divide the dataset into two non-overlapping sets, the training set ($80\%$) and the testing part ($20\%$). For cross-dataset evaluations, we use all $55,134$ images for testing.

\subsection{Creating the IMDB-Clean Dataset}\label{sec:creating-imdb-clean}
Although there have been efforts such as those reported in ~\cite{antipovApparentAgeEstimation2016, zhangAgeGroupGender2017} to manually clean the IMDB-WIKI dataset, many images still have incorrect annotations. This is mainly because the previous efforts either relied on simple heuristics to remove low-quality images~\cite{antipovApparentAgeEstimation2016}, or asked human raters to annotate apparent ages for the images based on their visual perception~\cite{zhangAgeGroupGender2017}. The latter is a very difficult task, resulting in incorrect guesses due to low quality images and very high quality make-ups.

To identify the source of noise, we revisited the annotation process for the images in the IMDB subset~\cite{rotheDEXDeepEXpectation2015}. 
We concluded that a relatively weak face detector was used to provide bounding box labels and that, when multiple faces are encountered, the one with the highest detection score is selected.

The main problem with such an annotation process is that when there are multiple faces, the adopted face \cite{Mathias2014facedetection} is biased towards large, frontal, middle-aged faces and give high scores to them. Another problem is that the utilised face detector fails to detect faces when the image has large variations in imaging quality, lightning, background~\etc, because it has not been trained on in-the-wild images. Some errors are shown in \figurename~\ref{fig:compare-imdb-clean}.

Based on the above analysis, we cleaned the dataset following the process below: 
\begin{enumerate}
    \item For each subject, we use an advanced face detector S$^3$FD~\cite{ZhangS3FD2017} to detect all faces in all images of the target subject crawled from IMDB. 
    \item We use FAN-Face~\cite{Yang_Bulat_Tzimiropoulos_2020} to map these face images into the face recognition embedding space.
    \item We then use a constrained version of the DBSCAN~\cite{dbscan2017} clustering algorithm to cluster these faces. Here, cannot-link constraints are applied to faces occurring in the same images. 
    \item Because the method can yield different results when the order of the input faces is changed, we repeat the clustering process multiple times using random ordering.
    \item After that, for each subject, we take the largest cluster obtained from all runs, and consider this to be the correct cluster containing the face images of the target subject. 
    \item For one subject, if the second largest cluster is larger than $70\%$ of the largest cluster, we consider this an ambiguous case. These ambiguous cases ($528$) are manually checked and filtered.
    \item Finally, we manually examine the dataset again to remove obvious mistakes caused by incorrect timestamps. 
\end{enumerate}

\figurename~\ref{fig:compare-imdb-clean} shows some noisy examples and the cleaned results. Note that the above cleaning process is not applied to the WIKI subset because most identities in this subset have only one image crawled from their Wikipedia page. 

\begin{figure}[ht]
    \centering
    \includegraphics[width=0.8\columnwidth]{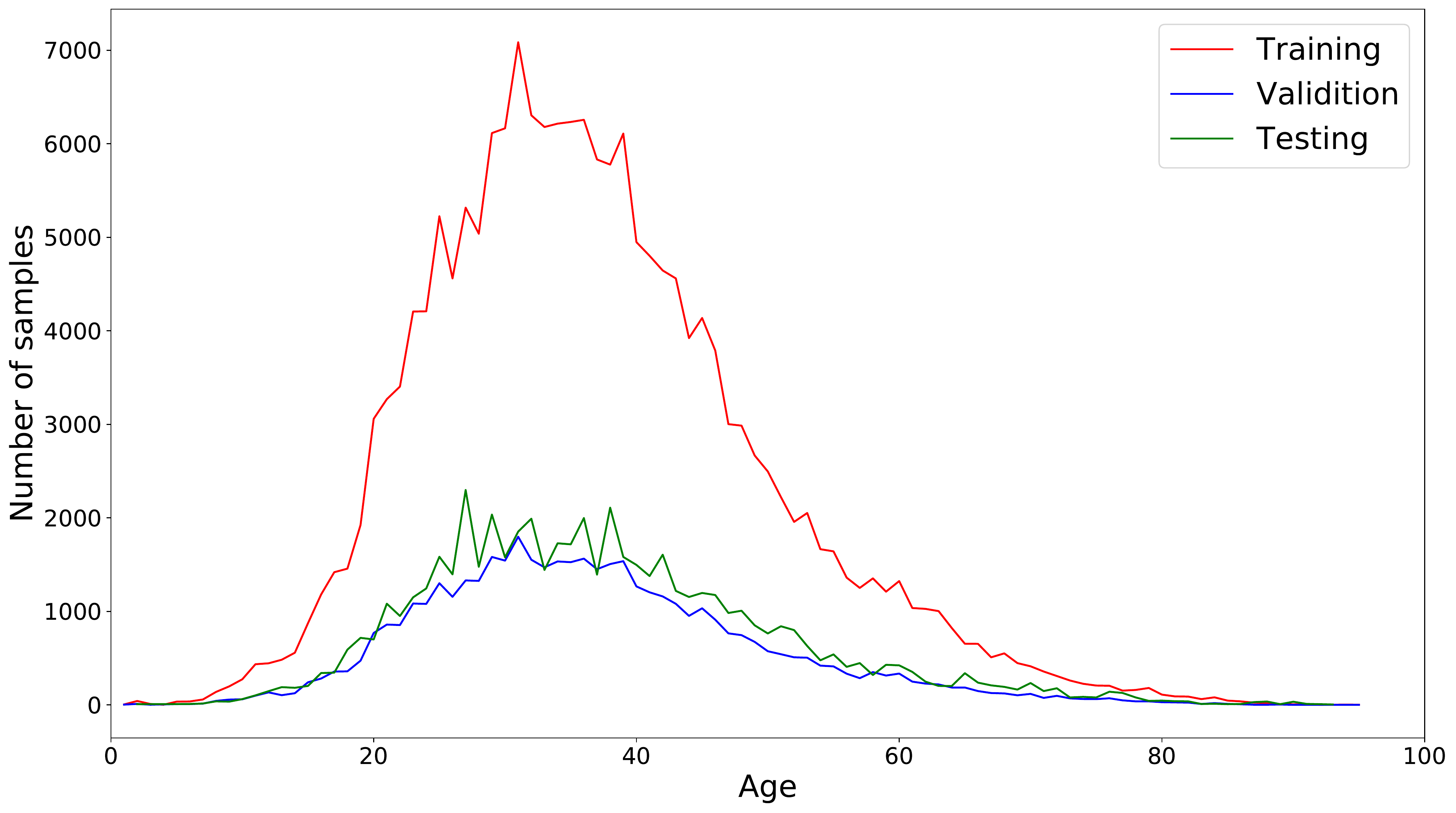}
    \caption{Age distributions of the proposed IMDB-Clean.}
    \label{fig:imdb-clean-dist}
\end{figure}

We refer to the cleaned dataset as IMDB-Clean, which contains $287,683$ images of $7,046$ subjects with age labels ranging from $0$ to $97$. We split IMDB-Clean into three subject-independent sets: training, validation and testing. The distributions of these sets are shown in \figurename~\ref{fig:imdb-clean-dist} and a comparison to other publicly available age datasets is given in Table~\ref{tab:age-datasets}.
\begin{table}[ht]
	\caption{Comparison of age estimation datasets used.}\label{tab:age-datasets}
	\begin{center}
		\begin{tabular}{lcccc}
    		\toprule
			Dataset & \# Images &\# ID & Age & In-the-wild? \\ \hline
			FG-Net~\cite{fgnet2002} & $1,002$ & $82$ & 0-69 & Yes \\ 	Morph~\cite{ricanekMORPHLongitudinalImage2006} & $55,134$ & $13,618$ & 16-77 & No \\ 
			CACD~\cite{chenFaceRecognitionRetrieval2015} & $163,446$ & $2,000$ & 14-62 & Yes \\ 
			KANFace~\cite{georgopoulosInvestigatingBiasDeep2020} & $41,036$ & $1,045$ & 0-100 & Yes \\ \hline
			IMDB-Clean~(ours) & ${287,683}$ & ${7,046}$ & 0-97 & Yes \\ \bottomrule
 			
		\end{tabular}
	\end{center}
\end{table}

\subsection{Evaluation Metrics}
The performance of models is measured by Mean Absolute Error (MAE) and Cumulative Score (CS). MAE is calculated using the average of the absolute errors between age predictions and groundtruth labels on the testing set; CS is calculated by CS$_l =\frac{N_l}{N}\cdot 100\%$ where $N$ is the total number of testing examples and $N_l$ is the number of examples whose absolute error between the estimated age and the groundtruth age is not greater than $l$ years. We report MAEs and CS$_5$ for all models.

\subsection{Implementation Details}
We use RoI Tanh-polar transform~\cite{linRoITanhpolarTransformer2021} to warp each input image to a Tanh-polar representation of resolution $512\times 512$. In the training stage, we apply image augmentation techniques including horizontal flipping, scaling, rotation and translation, as well as bounding box augmentations~\cite{linRoITanhpolarTransformer2021}.
For all experiments, we employed mini-batch SGD optimiser. The batch size, the weight decay and the momentum were set to $80$, $0.0005$ and $0.9$, respectively. The initial learning rate is 0.0001 and gradually increases to $0.01$ in $5$ epochs. Then the learning rate decreases exponentially at each epoch and the training is stopped either when the MAE on the validation set stops decreasing for $10$ epochs or we reach 90 training epochs. During testing, the test image and its flipped copy are fed into the model and their predictions are averaged.

For the comparisons reasons, we have re-implemented the following models from scratch: Dex~\cite{rotheDEXDeepEXpectation2015, rotheDeepExpectationReal2018}, OR-CNN~\cite{niuOrdinalRegressionMultiple2016}, DLDL~\cite{gaoDeepLabelDistribution2017}, DLDL-V2~\cite{gaoAgeEstimationUsing2018} and MV-Loss~\cite{panMeanVarianceLossDeep2018} while ResNet-18~\cite{resnet} was used as the backbone. The pre-processing, training and testing steps follow the above procedure. 
For the models with open-sourced training code, \ie C3AE~\cite{zhangC3AEExploringLimits2019}, SVRT~\cite{imScaleVaryingTripletRanking2019}, SSRNet~\cite{ssrnet2018} and Coral~\cite{Cao2020coral}, we used their default training setups and hyper-parameters. 
RetinaFace~\cite{Deng_2020_CVPR} was applied to detect $5$ facial landmarks (left and right eye centres, nose tip, left and right mouth corners). 
The input images were aligned using these landmarks with the method proposed in SSRNet\footnote{\url{https://github.com/shamangary/SSR-Net/blob/master/data/TYY_MORPH_create_db.py}} and then resized to $256\times 256$ pixels.

\section{Experiments}
\subsection{Can Face Parsing Mask Help?}\label{sec:exp:masks}
As a motivational example, we first test whether existing age estimation methods can benefit from facial parts segmentation. This is done by simply stacking the face parsing masks to the input image and using the resulted 14-channel tensor as the input to the models. During this experiment, we re-train three state-of-the-art methods, Dex, DLDL-V2 and MV-Loss, with the modified 14-channel input and test the models on IMDB-Clean. 
From Table~\ref{tab:stack-helps} we observe that by taking the stacked representation as input, all three models can achieve better performance in terms of both MAE and CS$_5$.
\begin{table}[h]
\renewcommand{\arraystretch}{1.3}
    \caption{Stacking images and face masks helps (evaluated on IMDB-Clean).}
    \label{tab:stack-helps}
    \centering
    \begin{tabular}{l|c|c}
    \toprule
        Method & MAE $\downarrow$ & CS$_5$(\%) $\uparrow$\\ \hline
        Dex~\cite{rotheDEXDeepEXpectation2015} &5.34 & 58.31 \\ 
        Dex with stacked input &  5.29 & 58.61\\\hline
        DLDL-V2~\cite{gaoAgeEstimationUsing2018} & 5.19  & 54.28 \\
        DLDL-V2 with stacked input & 5.12 & 55.14 \\\hline
        MV-Loss~\cite{panMeanVarianceLossDeep2018} & 5.27 & 53.97 \\
        MV-Loss with stacked input & 5.13 & 59.74 \\
    \bottomrule
    \end{tabular}
\end{table}

\subsection{Which Face Parsing Features to Use?}
We study which face parsing features are more informative for age estimation. 
We remove the face parsing attention module in FP-Age and use take face parsing features directly as input. 
We use four kinds of features as input: 1) low-level; 2) high-level; 3) stacking low and high; and 4) stacking low, high and mask.

From Table~\ref{tab:which-features}, we observe that using high-level features gives worse performance than using low-level features. This is consistent with earlier research~\cite{hanDemographicEstimationFace2015} which argues that local features are more informative as they capture ageing patterns around the facial regions, such as the dropping skin around the eyes, and the wrinkles around the mouth. On the other hand, due to the dilated convolutions in RTNet, the high-level features capture a larger perceptive field and thus the details can be lost. Stacking low-level and high-level features gives better performance which shows that these two types of features are complementary and combining them can help age estimation network. 

We also observe that adding mask further improves the model. This can be attributed to the fact that face mask contains semantics about different regions and adding it as an explicit attention mechanism helps the model to effortlessly locate these regions and extract ageing patterns. Furthermore, our face parsing attention module yields better results than simple stacking, which we further investigate in \sectionautorefname~\ref{sec:FPA}.
\begin{table}
    \caption{Using Different Face Parsing Features for Age Estimation on \imdbc.}
    \label{tab:which-features}
    \centering
    \begin{tabular}{l|c|c}
    \toprule
        Features from RTNet & MAE $\downarrow$ & CS$_5$(\%) $\uparrow$ \\ \hline
        Low-level &5.01 & 60.97 \\ 
        High-level &5.24 & 58.30 \\ 
        Stacking Low and High &4.96 & 61.01 \\ 
        Stacking Low, High and Masks&4.90 & 61.84 \\ \hline
        Full Model (with Face Parsing Attention) &4.68  &63.78 \\  
    \bottomrule
    \end{tabular}
\end{table}

\subsection{How about Other Feature Extractors?}
To validate the choice of the face parsing network as the feature extractor, we replace it with other CNN-based feature extractors and compare the performance of these variants. 

We adopted various generic feature extractors that are commonly used in transfer learning tasks as a replacement of face parsing network. The feature extractors include variants from the families of ResNet~\cite{resnet}, ResNeXt~\cite{xie2017resnext}, MobileNetV3~\cite{howard2019mobilenetv3}, FBNet~\cite{wu2019fbnet} and InceptionV4~\cite{szegedy2017inceptionv4}. Their weights have been pre-trained on the ImageNet dataset and remained frozen during the training for age estimation.

We also adopted a state-of-the-art face recognition feature network, ArcFace~\cite{deng2018arcface}, for feature extraction. The backbone of ArcFace is a customised, improved version of ResNet and it has been pre-trained on the large scale MS1M~\cite{guo2016ms1m} dataset for face recognition. The pre-trained weights remained frozen during the training for age estimation.

To ensure fair comparison, we did not use Face Parsing Attention in our model. All feature extractors adopted the same strategy for stacking deep and shallow semantic features. The age estimation sub-network and all other hyper-parameters remain the same as FPAge.

\tableautorefname~\ref{tab:which-feature-extractor} shows the results of using different pre-trained feature extractors on \imdbc. Our first observation is that the features of ResNet50 performed the best among ImageNet pre-trained models though being less accurate on image classification tasks. Moreover, all ImageNet pre-trained models obtained MAEs larger than 7. This in turn suggests generic features encoded in the CNNs for image classification are not directly transferable to solve the age estimation problem. 

Our second observation is that
face recognition features resulted better performance than generic features, meaning that the encoded details for distinguishing between identities are more transferable to the age estimation problem.

Finally, face parsing features have given the best performance among all evaluated backbones. This suggests that face parsing networks, designed to classify each pixel in a face, are able to encode the most informative details for age estimating.

\begin{table}[h]
    \caption{Performance of Different Pre-trained Feature Extractors on \imdbc.}
    \label{tab:which-feature-extractor}
    \centering
    \begin{tabular}{l|c|c|c|c}
    \toprule
        Feature extractor & Pre-train Data & \# Params & MAE $\downarrow$ & CS$_5$(\%) $\uparrow$ \\ \hline
        FBNet-C & ImageNet  & \textbf{2.9} M & 7.45 & 42.64 \\
        InceptionV4& ImageNet  & 41.1 M & 7.43 & 42.59 \\
        ResNeXt50& ImageNet & 23.0 M & 7.26 & 44.13 \\
        MobileNetv3-L& ImageNet  & 3.0 M & 7.24 & 44.11 \\
        ResNeXt101& ImageNet & 86.7 M & 7.17 & 44.28 \\ 
        ResNet101& ImageNet & 42.5 M & 7.12 & 44.63 \\
        ResNet50& ImageNet & 23.5 M & 7.10 & 44.59 \\
        ArcFace~\cite{deng2018arcface}& MS1M~\cite{guo2016ms1m} & 30.7 M & 5.96 & 52.19 \\ \hline
        Ours (w/o FPA) & iBugMask~\cite{linRoITanhpolarTransformer2021} & 27.3 M & 4.96 & 61.01 \\ 
        Ours (full) & iBugMask~\cite{linRoITanhpolarTransformer2021} & 27.3 M & \textbf{4.68} & \textbf{63.78} \\ 
    \bottomrule
    \end{tabular}
\end{table}

\subsection{But Aren't There Other Attentions?}
To validate the usefulness of the proposed Face Parsing Attention~(FPA) module, we compare it with three generic CNN attention modules, Squeeze-Excitation (SE)~\cite{Hu_2018_CVPR}, Convolutional Block Attention Module (CBAM)~\cite{Woo_2018_ECCV}, and Simple and Parameter Free Attention Module~(SimAM)~\cite{SimAM}. 
To ensure fair comparison, all attention modules are applied on the high-level features. Other components and all other hyper-parameters remain the same as FPAge.

\tableautorefname~\ref{tab:which-attention-module} shows that, when applied on the same face parsing features, the proposed FPA has achieved the lowest MAE on \imdbc. Moreover, FPA is directly derived from the face parsing map and acts as a probe to understand what the netword has learned, which we investigate in \sectionautorefname~\ref{sec:FPA}. 

\begin{table}[h]
    \caption{Applying Different Attention Modules on Face Parsing Features on \imdbc.}
    \label{tab:which-attention-module}
    \centering
    \begin{tabular}{l|c|c}
    \toprule
        Attention & MAE $\downarrow$ & CS$_5$(\%) $\uparrow$ \\ \hline
        Squeeze-Excitation~\cite{Hu_2018_CVPR} & 4.86 & 61.47 \\
        CBAM\cite{Woo_2018_ECCV} & 4.83 & 62.04 \\
        SimAM\cite{SimAM} & 4.82 & 62.03 \\
        \hline
        Face Parsing Attention~(ours) & \textbf{4.68} & \textbf{63.78} \\
    \bottomrule
    \end{tabular}
\end{table}

\subsection{Ablation Study}

We conduct ablation study on the overall \fpage~model to understand the contribution of each component. We have evaluated five variants on \imdbc.  We replaced the face parsing network with ResNet50 which has the similar number of parameters. Next, we either removed the FPA module or replace it with the Squeeze-Excitation~\cite{Hu_2018_CVPR} module. 

Table~\ref{tab:ablation-study} shows that the biggest improvement comes from adopting the face parsing network as the feature extractor with MAEs improved from above 7 to 4.96. Moreover, the proposed FPA has reduced the MAE from 4.96 to 4.68, an improvement of 0.28 compared with 0.1 by the Squeeze-Excitation module.

\begin{table}[h]
    \caption{Ablation Study on \imdbc.}
    \label{tab:ablation-study}
    \centering
    \begin{tabular}{c|c|c|c}
    \toprule
        Feature Extractor & Attention & MAE $\downarrow$ & CS$_5$(\%) $\uparrow$ \\ \hline
        ResNet50 & - & 7.10 & 44.59 \\
        ResNet50 & Squeeze-Excitation & 7.00 & 45.50 \\
        Face Parsing Network & - & 4.96 & 61.01 \\
        Face Parsing Network & Squeeze-Excitation &4.86 & 61.47\\ \hline
        Face Parsing Network  & FPA & \textbf{4.68} & \textbf{63.78}\\
        
    \bottomrule
    \end{tabular}
\end{table}
\subsection{What Did FPA Learn Really?}\label{sec:FPA}
\begin{figure}
    \centering
    \includegraphics[width=0.8\columnwidth]{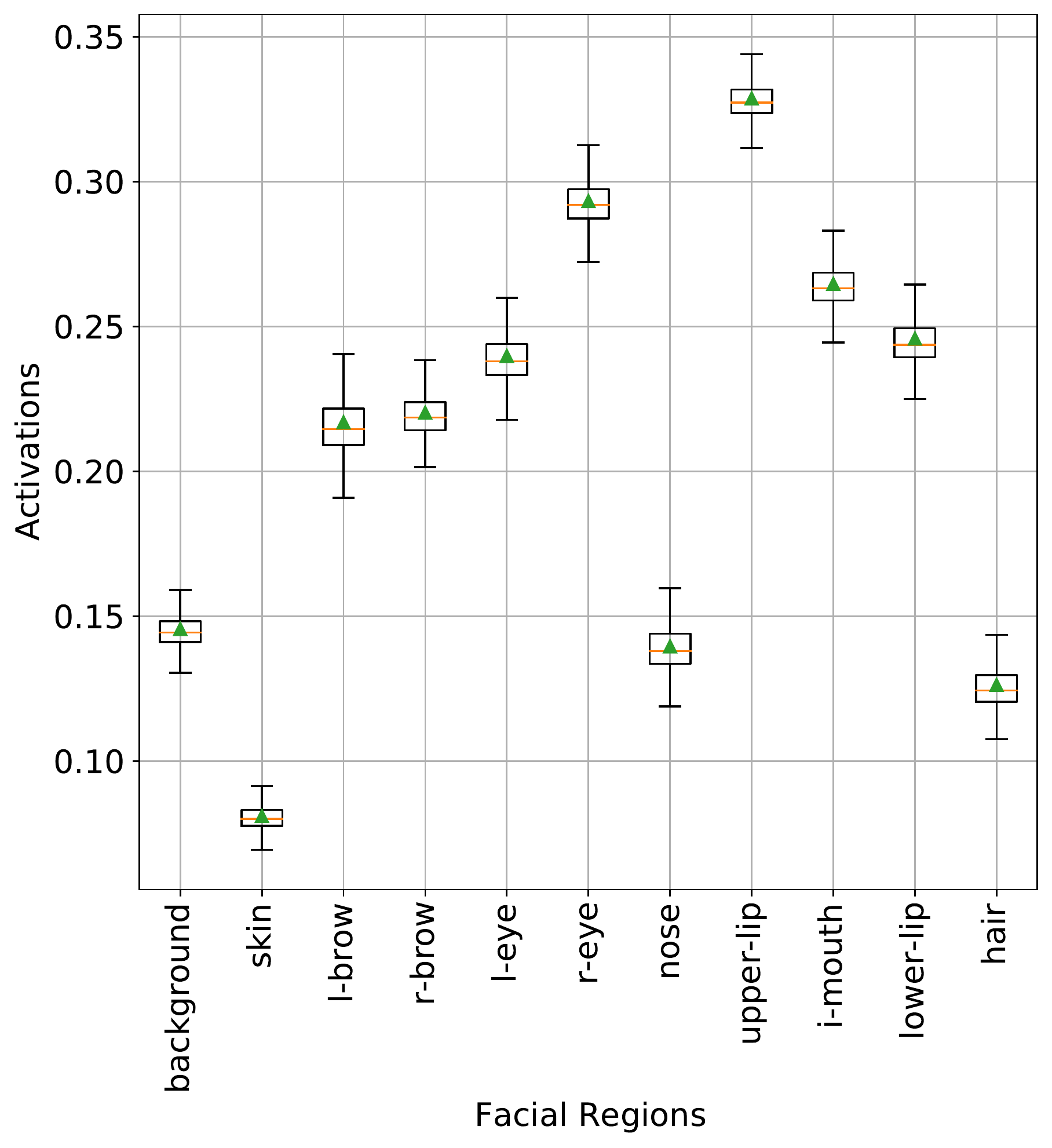}
    \caption{Attention weights for facial regions induced by the face parsing attention module on IMDB-Clean.}
    \label{fig:se_act}
\end{figure}

To provide a clearer picture of the function of the proposed face parsing attention module, we study the 11-class activation output of the Sigmoid layer. Specifically we show the mean and standard deviations of the activations for images in the IMDB-Clean dataset in \figurename~\ref{fig:se_act}. 

We observe that the network consistently gives higher attention weights to most inner facial regions, especially eyes (``l-eye'' and ``r-eye'') and mouth (``upper-lip'', ``i-mouth'', and ``lower-lip''). This is in line with the observations reported in \cite{hanDemographicEstimationFace2015}.
Interestingly, it can also be seen that the ``background`` class contributes more than the ``skin'' class. This could be attributed to the fact that the face parsing network classifies objects like ``beard'', ``glasses'' and ``accessories'' as ``background'', and such context information could give hints about the person's age.

We have also performed the same test on separate age groups and observed the importance of different facial regions follows the same trend as shown in \figurename~\ref{fig:se_act}. This means that the face parsing attention allows the model to focus on informative regions that are universally important for judging different ages.
Although there are some works such as ~\cite{yiAgeEstimationMultiscale2014,Angeloni_2019_ICCV, liao2017local, peiAttendedEndtoEndArchitecture2020}  that used attention, we are the first to present the evidence that the network attends to specific facial parts and that such attention modelling improves age estimation.

\subsection{Effectiveness of IMDB-Clean}
We conduct experiments on the effectiveness of the proposed IMDB-Clean. Specifically, we train 6 models on three datasets, \ie~IMDB-Clean, IMDB-WIKI and CACD. We then directly test them on KANFace without any fine-tuning. For IMDB-WIKI, we randomly sampled $300,000$ images for training; for the other two datasets, we used their provided training splits.
Table~\ref{tab:imdb-clean-effectiveness} shows the cross-dataset evaluation results on KANFace. We observe that 1) all models have improved when they are trained on our IMDB-Clean; 2) our model outperforms other methods when trained on IMDB-Clean and IMDB-WIKI, and is comparable to DLDL-V2 when trained on CACD. 
\begin{table}[ht]
	\caption{Effectiveness of IMDB-Clean (Testing dataset: KANFace~\cite{georgopoulosInvestigatingBiasDeep2020}). }\label{tab:imdb-clean-effectiveness}
	\begin{center}
	\begin{threeparttable}
	\begin{adjustbox}{width=\columnwidth}
		\begin{tabular}{l|c|c|c|c|c|c}
    		\toprule
			 \multirow{2}{*}{}& \multicolumn{6}{c}{Trained on}  \\ 
			& \multicolumn{2}{c|}{IMDB-Clean} & \multicolumn{2}{c|}{IMDB-WIKI} & \multicolumn{2}{c}{CACD}
			\\ 
			\midrule \hline
            Method & MAE & CS$_5$(\%) & MAE & CS$_5$(\%) & MAE & CS$_5$(\%) 
            \\ 
            \hline 
			DLDL\cite{gaoDeepLabelDistribution2017}  & \textbf{9.84} &\textbf{37.37}   & 12.19 & 27.20 &11.66 &29.20  \\
			DLDL-V2\cite{gaoAgeEstimationUsing2018}   &\textbf{8.05} & \textbf{41.74} & 11.46 & 28.83 &10.88 & 30.66   \\
			Dex\cite{rotheDeepExpectationReal2018}   & \textbf{7.91} & \textbf{42.30} & 11.70 & 20.91 & 11.90 & 28.62   \\
			M-V Loss\cite{panMeanVarianceLossDeep2018}   & \textbf{7.71} & \textbf{43.31}& 11.95 & 28.30  & 11.30&   29.07 \\
			OR-CNN\cite{niuOrdinalRegressionMultiple2016}   & \textbf{7.71} & \textbf{47.51} & 11.10 & 33.07 & 11.18  & 32.90 \\
			\hline
			FP-Age~(ours)   & \textbf{6.81} & \textbf{48.49} & 10.83 & 29.63 & 10.91 &  30.27  \\
			\bottomrule
		\end{tabular}
		\end{adjustbox}
	\end{threeparttable}
	\end{center}
\end{table}

\subsection{Comparison to the State-of-the-arts} \label{sec:compare-sotas}
\subsubsection{Intra-Dataset Evaluation}
In this section, the performance of the proposed FP-Age is compared with the state-of-the-art age estimation methods under the intra-dataset evaluation protocol. Three benchmarks are used: IMDB-Clean, Morph and CACD. On IMDB-Clean, we train all the models from scratch on the same training set and test them on the testing set. For Morph and CACD, we only train our own models and compare the performance with the reported values for the other methods on the testing set.

The benchmarking results are shown in Table~\ref{tab:imdb-clean-results}. It can be seen that our model achieves state-of-the-art results on \imdbc~dataset. When all model are trained under the same settings, our model achieves $4.68$ in terms of MAE and $63.78\%$ in terms of CS$_5$. Additionally, the results show that \imdbc~is quite challenging compared to other datasets, such as Morph where the state-of-the-art MAEs have achieved below $2$. We provide significance testing analysis in Appendix~\ref{appendix:t-test} which shows our results are significantly better than the other methods.

From Table~\ref{tab:morph-results-s3}, it can be seen that our model achieves state-of-the-art results on Morph dataset. When directly trained on Morph, our model achieves $2.04$ in terms of MAE and $92.8\%$ in terms of CS$_5$. When pre-trained on IMDB-Clean and fine-tuned the weights on Morph, \fpage~achieves a MAE of $1.90$ and a  CS$_5$ of $93.7\%$, which is the new state-of-the-art result.

Table~\ref{tab:cacd-results} shows the results on the CACD dataset. Following the training protocols of CACD~\cite{shenDeepRegressionForests2018}, we train our models with both the training set and the validation set, and report the MAE values on the testing set. Our model achieves $4.50$ when trained on CACD-train and $5.62$ when trained on CACD-val. Similar to above experiments, when pre-trained on IMDB-Clean, our model achieves $4.33$ and $4.95$ .  
\subsubsection{Cross-Dataset Evaluation} To test the generalisation ability of different models, we conduct experiments on a cross-dataset evaluation protocol. Our results are compared with $9$ advanced models: SSRNet, C3AE, SVRT, DLDL, DLDL-V2, Coral, Dex, MV-Loss, and OR-CNN. We train all models on IMDB-Clean and test them on $4$ different testing datasets without fine-tuning. The reuslts are summarised in Table~\ref{tab:cross-db-res}. It can be seen that when all models are trained on \imdbc, the proposed \fpage~achieves the best results on most of evaluation datasets.
\begin{table}[tb] 
	\caption{Intra-Dataset Evaluation on \imdbc.}\label{tab:imdb-clean-results}
	\begin{center}
    	\begin{threeparttable}
    		\begin{tabular}{l|c|c|c}
    			\toprule
    			{Method} & {MAE} $\downarrow$  & {CS$_5$(\%)} $\uparrow$ & {Year}\\
    			\midrule
    			\midrule
    	    	OR-CNN~\cite{niuOrdinalRegressionMultiple2016} & 5.85
& 49.72 & 2016 \\		
    			DLDL~\cite{gaoDeepLabelDistribution2017}  & 6.04 & 56.94 & 2017 \\
    			SSRNet~\cite{ssrnet2018} &7.08 & 27.87 & 2018 \\
    			Dex~\cite{rotheDeepExpectationReal2018}  & 5.34 & 58.61 & 2018 \\
    			M-V Loss\cite{panMeanVarianceLossDeep2018} & 5.27 & \textit{59.74} & 2018 \\
    		    DLDL-V2~\cite{gaoAgeEstimationUsing2018}  & \textit{5.19} & 54.28 & 2018 \\
    			SVRT~\cite{imScaleVaryingTripletRanking2019} & 5.85 & 49.72 & 2019 \\
    			C3AE~\cite{zhangC3AEExploringLimits2019} &6.75 & 47.98 & 2019 \\
    			\midrule
    			FP-Age~(ours) & \textbf{4.68}\tnote{$\ddagger$} & \textbf{63.78} &{-} \\
    			\bottomrule
    		\end{tabular}

    	\begin{tablenotes}
    	\item \textbf{Bold} indicates the best and \textit{italic} the second
    	\item $\ddagger$ Our results are statistically significant according to paired t-test and Bonferroni corrections (See Appendix~\ref{appendix:t-test})
    	\end{tablenotes}
    	\end{threeparttable}
	\end{center}
\end{table}

\begin{table}[tb] 
	\caption{Intra-Dataset Evaluation on Morph~\cite{ricanekMORPHLongitudinalImage2006}.}\label{tab:morph-results-s3}
	\begin{center}
    	\begin{threeparttable}
    		\begin{tabular}{l|c|c|c}
    			\toprule
    			{Method} & {MAE} $\downarrow$  & {CS$_5$(\%)} $\uparrow$&{Year} \\
    			\midrule
    			\midrule
    			Human workers~\cite{niuOrdinalRegressionMultiple2016} & 6.30 & 51.0  & 2015 \\
    	    	OR-CNN~\cite{niuOrdinalRegressionMultiple2016} & 3.34 & 81.5 & 2016 \\		
    			DLDL~\cite{gaoDeepLabelDistribution2017}  & 2.42 & - & 2017 \\
    			ARN~\cite{agustssonAnchoredRegressionNetworks2017} & 3.00 & - & 2017 \\	    			Ranking-CNN~\cite{chenUsingRankingCNNAge2017}\tnote{$\ast$}  & 2.96 & 85.2 & 2017 \\
    			M-V Loss\cite{panMeanVarianceLossDeep2018} & 2.41 & 91.2 & 2018 \\
    			DLDL-V2~\cite{gaoAgeEstimationUsing2018}\tnote{$\dagger$}  & {1.97} & - & 2018 \\
    			BridgeNet~\cite{liBridgeNetContinuityAwareProbabilistic2019}\tnote{$\ast$} & 2.38 & - & 2019 \\
    			C3AE~\cite{zhangC3AEExploringLimits2019}\tnote{$\ast$} & 2.75 & - & 2019 \\
    			AVDL~\cite{wenAdaptiveVarianceBased2020}\tnote{$\ast$} & \textit{1.94} & - & 2020 \\
    			PML~\cite{dengPMLProgressiveMargin2021} & {2.15} & -  &  {2021} \\ 
    			DRF~\cite{shenDeepDifferentiableRandom2021} & 2.14 & 91.3 & 2021 \\ 
    			\midrule
    			FP-Age~(ours) & 2.04& \textit{92.8} &{-} \\
    			FP-Age\tnote{$\ddagger$}~~(ours) & \textbf{1.90} & \textbf{93.7} &{-} \\
    			\bottomrule
    		\end{tabular}

    	\begin{tablenotes}
    	\item \textbf{Bold} indicates the best and \textit{italic} the second
    	\item $^\ast$ pre-trained on IMDB-WIKI
    	\item $^\dagger$ pre-trained on MS-Celeb-1M
    	\item $\ddagger$ pre-trained on the proposed IMDB-Clean 
    	\end{tablenotes}
    	\end{threeparttable}
	\end{center}
\end{table}

\begin{table}[tb]
	\caption{Intra-Dataset Evaluation (MAEs) on CACD~\cite{chenFaceRecognitionRetrieval2015}.}\label{tab:cacd-results}
	\begin{center}
	\begin{threeparttable}
		\begin{tabular}{l|c|c|c}
    		\toprule
			\multirow{2}{*}{Method} & \multicolumn{2}{c|}{Trained on} &\multirow{2}{*}{Year} \\
			& CACD-train & CACD-val & \\
			\midrule \midrule
			Dex\cite{rotheDeepExpectationReal2018} & 4.78 & 6.52 & 2018 \\
			DLDLF~\cite{shenDeepRegressionForests2018} & 4.67 & 6.16 & 2018 \\
			DRF~\cite{shenDeepDifferentiableRandom2021} &4.61 & 5.63 & 2021 \\
			\hline
			FP-Age~(ours) & 4.50 & 5.62 & -\\
			FP-Age\tnote{$\ddagger$}~~(ours) & \textbf{4.33} &  \textbf{4.95} & - \\ \bottomrule
		\end{tabular}
    \begin{tablenotes}
    \item $\ddagger$ pre-trained on the proposed IMDB-Clean 
    \end{tablenotes}
	\end{threeparttable}
	\end{center}
\end{table}

\begin{table*}[ht] 
	\caption{Cross-Dataset Evaluation (Training set: IMDB-Clean).}\label{tab:cross-db-res}
	\begin{center}
    		\begin{threeparttable}
		\begin{tabular}{l|c|c|c|c|c|c|c|c}
    		\toprule
            & \multicolumn{2}{c|}{FG-Net~\cite{fgnet2002}} 
            & \multicolumn{2}{c|}{Morph~\cite{ricanekMORPHLongitudinalImage2006}} 
            &
            \multicolumn{2}{c|}{KANFace~\cite{georgopoulosInvestigatingBiasDeep2020}} & \multicolumn{2}{c}{CACD-test~\cite{chenFaceRecognitionRetrieval2015}}
			\\ 
			\midrule \hline
            Method & MAE & CS$_5$(\%) & MAE & CS$_5$(\%) & MAE & CS$_5$(\%) 
            & MAE & CS$_5$(\%) 
            \\ 
            \hline 
			SSRNet\tnote{$\ast$}~~\cite{ssrnet2018} & 12.04 & 19.86 & 7.12 & 40.77 & 11.36 & 30.11 & 11.76&22.01 
			\\
			C3AE\tnote{$\ast$}~~\cite{zhangC3AEExploringLimits2019} & 11.23 
			& 27.34 
			& 7.03 
			& 41.81
			& 10.41
			& 31.71
			& 12.71
			& 16.14
			\\
			SVRT\tnote{$\ast$}~~\cite{imScaleVaryingTripletRanking2019} & 9.77 & 23.75 & 5.87 &43.71 &10.89 & 27.55 & 11.73 & 14.37 
			\\
			DLDL\tnote{$\dagger$}~~\cite{gaoDeepLabelDistribution2017}  & 11.40 & 24.05 &6.07 & 33.06 & 9.84 &37.37 &6.53 &  55.12 
\\
			Coral\tnote{$\ast$}~~\cite{Cao2020coral}    
			& 6.12
			& 45.61
			& 6.13
			& 42.33
			& 7.88
			& 39.01
			& 12.58
			& 11.38
			\\
			Dex\tnote{$\dagger$}~~\cite{rotheDeepExpectationReal2018}  & 6.52 & 41.52 &  5.63 & 53.03 & 7.91 & 42.30 & 6.08 & 55.94 
			\\
			DLDL-V2\tnote{$\dagger$}~~\cite{gaoAgeEstimationUsing2018}   & 6.65 & 42.41  & 5.10 & 55.64 &8.05 & 41.74  & 5.92 & 57.39 
			\\
			M-V Loss\tnote{$\dagger$}~~\cite{panMeanVarianceLossDeep2018}   & 6.49  & 42.12  &4.99 &56.94 & 7.71 & 43.31 & 5.88 & 57.22 
			\\
			OR-CNN\tnote{$\dagger$}~~\cite{niuOrdinalRegressionMultiple2016} 
			& 6.44
			& 40.72
			&  5.04
			& 60.87
			& 7.71
			& 47.51
			& 5.83
			& \textbf{62.47}
			\\
			\hline
			Ours\tnote{$\dagger$}   & \textbf{5.60} &  \textbf{48.80} & \textbf{4.67} & \textbf{60.54} & \textbf{6.81} & \textbf{48.49} & \textbf{5.60} & 60.91 \\ \bottomrule

		\end{tabular}
    	\begin{tablenotes}
    	\item $^\ast$ inputs are pre-processed with 5-point face alignment 
    	\item $^\dagger$ inputs are pre-processed with RoI Tanh-polar Transform~\cite{linRoITanhpolarTransformer2021}
    	\end{tablenotes}
	\end{threeparttable}

	\end{center}
\end{table*}

\section{Conclusion}
In this paper, we have proposed a simple yet effective approach of exploiting face parsing semantics for age estimation. We have designed a framework to aggregate features from different levels of the face parsing network. A novel face parsing attention module is proposed to explicitly introduce facial semantics into the age estimation network. To train the model, we propose an semi-automatic clustering method for cleaning existing dataset and introduce the resulting IMDB-Clean dataset as a new in-the-wild benchmark.
Thanks to the attention mechanism and the large-scale dataset, we have observed that the network focuses on certain facial parts when predicting ages. The nose region appears least informative for age estimation. Moreover, the extensive experiments have shown that our model outperforms the current state-of-the-art methods on various dataset in both intra-dataset and cross-dataset evaluations.  To the best of our knowledge, this is the first attempt of leveraging face parsing attention to achieve age estimation. We hope our design could inspire the readers to consider similar attention models for different deep face analysis tasks. 

For future work, since we have identified that all models performed less favourably in the cross-dataset evaluation, an interesting direction would be to investigate domain shifts between different datasets and find out how to mitigate them. Also, as most works focus on image-based age estimation, it would also be interesting to extend the models to videos and study how to improve the them with temporal information from videos. We will explore these ideas in the future.

\appendix[Statistical Significance Analysis]\label{appendix:t-test}
We conduct paired t-tests on the Absolute Error (AE) on the testing set of \imdbc~between FP-Age and the other eight methods, \ie OR-CNN~\cite{niuOrdinalRegressionMultiple2016}, DLDL~\cite{gaoDeepLabelDistribution2017}, SSRNet~\cite{ssrnet2018}, Dex~\cite{rotheDeepExpectationReal2018}, MV-Loss~\cite{panMeanVarianceLossDeep2018}, DLDL-V2~\cite{gaoAgeEstimationUsing2018}, SVRT~\cite{imScaleVaryingTripletRanking2019} and C3AE~\cite{zhangC3AEExploringLimits2019}. Concretely, suppose there are $N$ images in the testing set, then $\epsilon^{\text{\fpage}}_i $ is the AE between the predicted age of FP-Age and the groundtruth age on the $i$-th testing image and $\epsilon^{M}_i$ is such AE for another method $M$. The difference between the $i$-th pair is defined as $d_i = \epsilon^{\text{\fpage}}_i - \epsilon^{M}_i$. The t statistic is calculated as 
\begin{equation}
    t = \sqrt{N} \frac{\Bar{d}}{\sigma_d}
\end{equation}
where $\Bar{d}$ and $\sigma_d$ are the average and standard deviation of $\{d_i\}_{i=1}^N$. We correct the p-values using Bonferroni correction. The alpha value is set to $0.05$. From Table~\ref{tab:imdb-clean-results} and Table~\ref{tab:t-test}, we observe our results are significantly better than those of the other methods. We can, thus, reject the null hypotheses.
\begin{table}[tb] 
	\caption{Paired t-Tests between FP-Age and Other Methods on \imdbc.}\label{tab:t-test}
	\begin{center}
    	\begin{threeparttable}
    		\begin{tabular}{l|c|c|c}
    			\toprule
    			{Method} & {t-statistic}  & {p-value} & {Corrected p-value}\\
    			\midrule
    			\midrule
    			SSRNet~\cite{ssrnet2018} & -137.73 & 0.00\tnote{$\ast$} & 0.00\tnote{$\ast$} \\
    			C3AE~\cite{zhangC3AEExploringLimits2019} &-84.66 & 0.00\tnote{$\ast$} & 0.00\tnote{$\ast$} \\
    		    DLDL~\cite{gaoAgeEstimationUsing2018}  & -66.44 & 0.00\tnote{$\ast$} & 0.00\tnote{$\ast$} \\
    			Dex~\cite{rotheDeepExpectationReal2018}  & -39.08 & 0.00\tnote{$\ast$} & 0.00\tnote{$\ast$} \\
    	    	OR-CNN~\cite{niuOrdinalRegressionMultiple2016} & -33.83 & \num{2.08e-248} & \num{1.17e-243}   \\
    			DLDL-V2~\cite{gaoAgeEstimationUsing2018} & -31.83 & \num{2.24e-220} & \num{1.26e-220} \\
    			M-V Loss\cite{panMeanVarianceLossDeep2018} & -28.03 & \num{1.21e-171} & \num{6.80e-167} \\
    			SVRT~\cite{imScaleVaryingTripletRanking2019} & -22.89 & \num{2.01e-115} & \num{1.23e-110} \\
    			\bottomrule
    		\end{tabular}

    	\begin{tablenotes}
    	\item $\ast$ indicates underflow
    	\end{tablenotes}
    	\end{threeparttable}
	\end{center}
\end{table}

\section*{Acknowledgment}
Data cleaning and all experiments have been conducted at Imperial College London.

\ifCLASSOPTIONcaptionsoff
  \newpage
\fi

\bibliographystyle{IEEEtran}
\bibliography{age}

\begin{IEEEbiography} [{\includegraphics[width=1in,height=1.25in,clip,keepaspectratio]{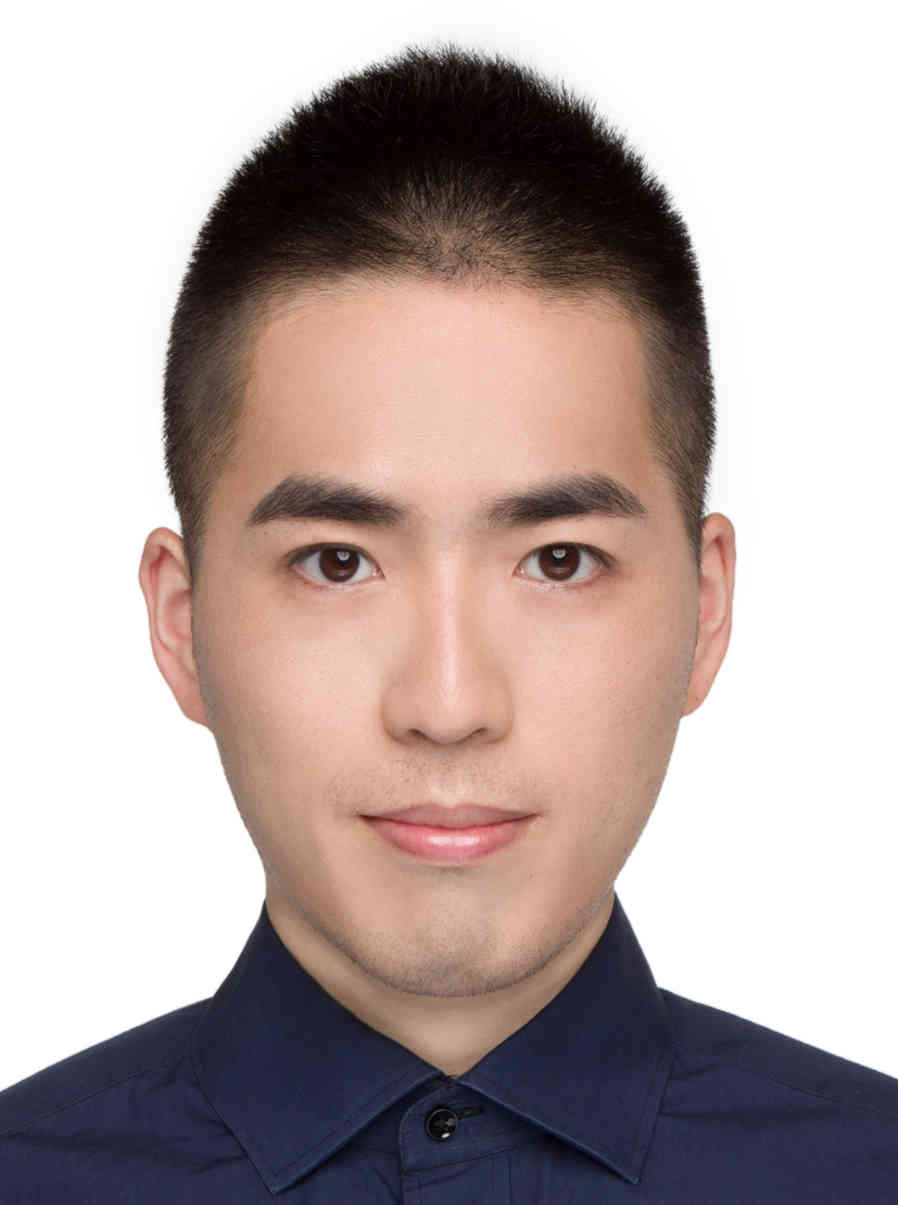}}]
{Yiming Lin} is a research scientist at Meta AI (formerly known as Facebook AI). He
received his PhD degree in 2021, and his MSc degree with Distinction in Communications and Signal Processing in 2016, from Imperial College London. His research interests include face tracking, face parsing and facial attribute recognition. He is a member of the IEEE.
\end{IEEEbiography}
\begin{IEEEbiography}
 [{\includegraphics[width=1in,height=1.25in,clip,keepaspectratio]{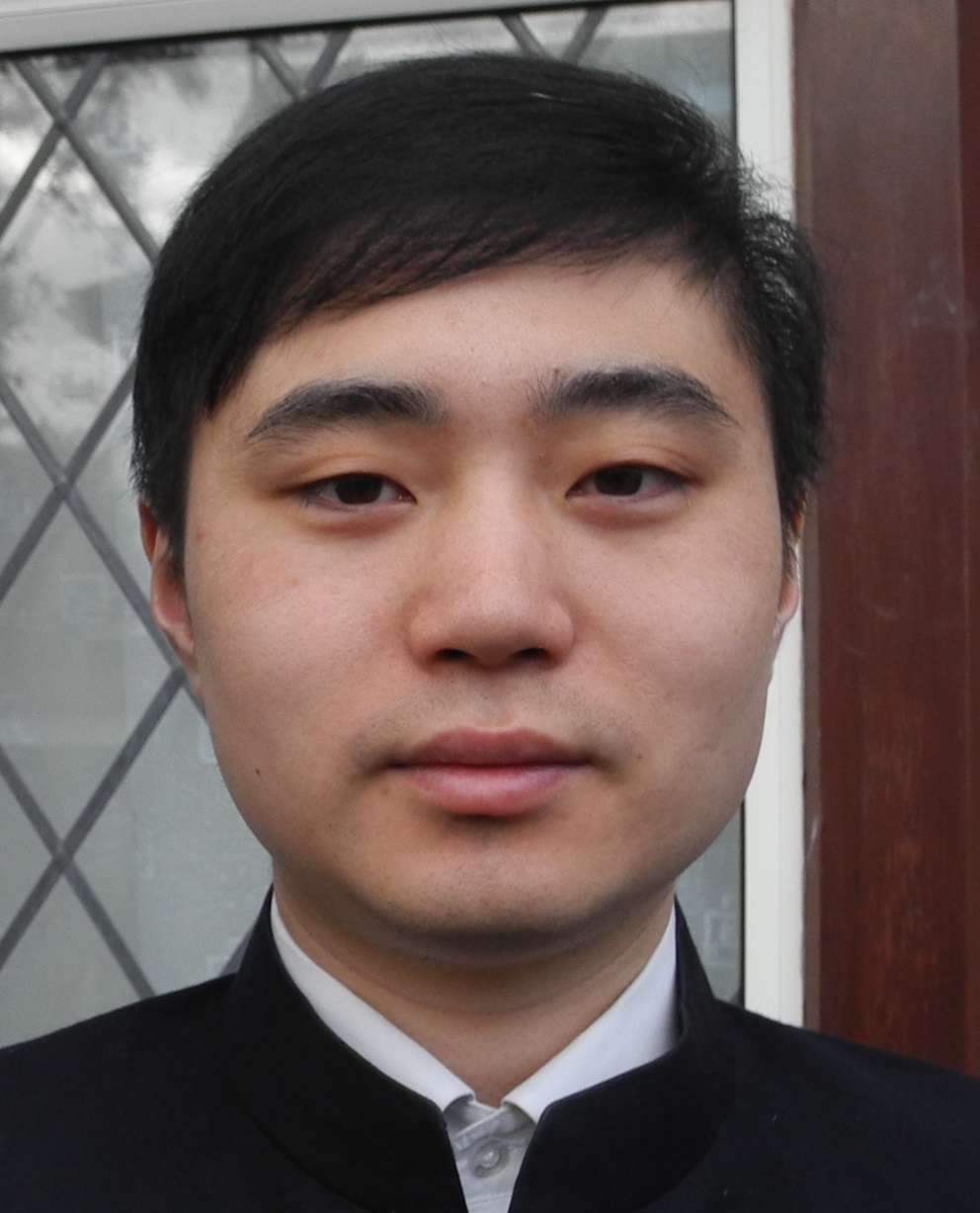}}]
 {Jie Shen} is a research scientist at Meta AI and an honorary research fellow at the Department of Computing at Imperial College London. He received his B.Eng. in electronic engineering from
Zhejiang University in 2005, and his MSc in advanced computing and Ph.D. from Imperial College London in 2008 and 2014. His research interests include facial analysis, computer vision, affective computing, and social robots. He is a member of the IEEE.
\end{IEEEbiography}

\begin{IEEEbiography}[{\includegraphics[width=1in,height=1.25in,clip,keepaspectratio]{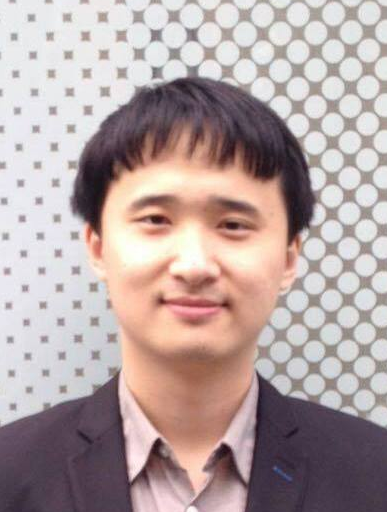}}]{Yujiang Wang}
is a postdoctoral researcher at the University of Oxford. He received his Ph.D. degree from Imperial College London in February 2021, after which he worked as a research collaborator at Meta AI until January 2022. He obtained a BSc degree in Architecture from Tsinghua University in 2010, and two MSc from University College London and Imperial College London, respectively. His research interest centres around video face parsing and clustering, word-level lip-reading, smart wearable devices, clinical AI, etc. 
\end{IEEEbiography}

\begin{IEEEbiography} 
 [{\includegraphics[width=1in,height=1.25in,clip,keepaspectratio]{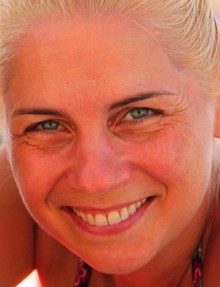}}]
 {Maja Pantic} is a professor in affective and behavioural computing in the Department of Computing at Imperial College London, UK. She was the Research Director of Samsung AI Centre, Cambridge, UK from 2018 to 2020 and is currently an AI Scientific Research Lead at Meta Platforms (Facebook) London. She currently serves as an associate editor for International Journal of Computer Vision. She has received various awards for her work on automatic analysis of human behaviour including the Royal Society Roger Needham Award 2011 and IAPR Maria Petrou Award 2020. She is a fellow of the UK’s Royal Academy of Engineering, the IEEE, and the IAPR.
\end{IEEEbiography}

\end{document}